\documentclass[10pt,twocolumn,letterpaper]{article}

\usepackage[pagenumbers]{cvpr} 

\usepackage{graphicx}
\usepackage{amsmath}
\usepackage{amssymb}
\usepackage{booktabs}

\usepackage{algorithm}
\usepackage{algorithmic}
\usepackage{amsthm}
\usepackage{paralist}

\usepackage{listings}
\usepackage{multirow}
\usepackage[accsupp]{axessibility}

\def \R {\mathcal{R}}
\def \V {\mathcal{V}}

\def \x {\mathbf{x}}
\def \cc {\mathbf{c}}
\def \y {\mathbf{y}}
\def \OO {\mathcal{O}}

\def \y {\mathbf{y}}

\def \p {\mathbf{p}}

\def \LL {\mathcal{L}}

\newtheorem{thm}{Theorem}

\newtheorem{cor}{Corollary}

\usepackage[pagebackref,breaklinks,colorlinks]{hyperref}

\usepackage[capitalize]{cleveref}
\crefname{section}{Sec.}{Secs.}
\Crefname{section}{Section}{Sections}
\Crefname{table}{Table}{Tables}
\crefname{table}{Tab.}{Tabs.}

\def\cvprPaperID{5628} 
\def\confName{CVPR}
\def\confYear{2022}

\begin{document}

\title{Unsupervised Visual Representation Learning by Online Constrained K-Means}

\author{Qi Qian$^1$\quad Yuanhong Xu$^2$\quad Juhua Hu$^3$\quad Hao Li$^2$\quad Rong Jin$^1$\\
$^1$Alibaba Group, Bellevue, WA 98004, USA\\
$^2$Alibaba Group, Hangzhou, China\\
$^3$School of Engineering and Technology, University of Washington, Tacoma, WA 98402, USA\\
{\tt\small \{qi.qian, yuanhong.xuyh, lihao.lh, jinrong.jr\}@alibaba-inc.com, juhuah@uw.edu}
}

\maketitle

\begin{abstract}
Cluster discrimination is an effective pretext task for unsupervised representation learning, which often consists of two phases: clustering and discrimination. Clustering is to assign each instance a pseudo label that will be used to learn representations in discrimination. The main challenge resides in clustering since prevalent clustering methods (e.g., k-means) have to run in a batch mode. Besides, there can be a trivial solution consisting of a dominating cluster. To address these challenges, we first investigate the objective of clustering-based representation learning. Based on this, we propose a novel clustering-based pretext task with online \textbf{Co}nstrained \textbf{K}-m\textbf{e}ans (\textbf{CoKe}). Compared with the balanced clustering that each cluster has exactly the same size, we only constrain the minimal size of each cluster to flexibly capture the inherent data structure. More importantly, our online assignment method has a theoretical guarantee to approach the global optimum. By decoupling clustering and discrimination, CoKe can achieve competitive performance when optimizing with only a single view from each instance. Extensive experiments on ImageNet and other benchmark data sets verify both the efficacy and efficiency of our proposal. Code is available at \url{https://github.com/idstcv/CoKe}.
\end{abstract}

\section{Introduction}

Recently, many research efforts have been devoted to unsupervised representation learning that aims to leverage the massive unlabeled data to obtain applicable models. Different from supervised learning, where labels can provide an explicit discrimination task for learning, designing an appropriate pretext task is essential for unsupervised representation learning. Many pretext tasks have been proposed, e.g., instance discrimination~\cite{DosovitskiyFSRB16}, cluster discrimination~\cite{CaronBJD18}, invariant mapping~\cite{ChenH21,GrillSATRBDPGAP20}, solving jigsaw puzzles~\cite{NorooziF16}, patch inpainting~\cite{PathakKDDE16}, etc. Among them, instance discrimination that identifies each instance as an individual class~\cite{DosovitskiyFSRB16} is popular due to its straightforward objective. However, this pretext task can be intractable on large-scale data sets. Consequently, contrastive learning is developed to mitigate the large-scale challenge~\cite{ChenK0H20,He0WXG20,WuXYL18} with a memory bank~\cite{He0WXG20} or training with a large mini-batch of instances~\cite{ChenK0H20}, which requires additional computation resources.

\begin{figure}[t]
\centering
\includegraphics[height = 1.5in]{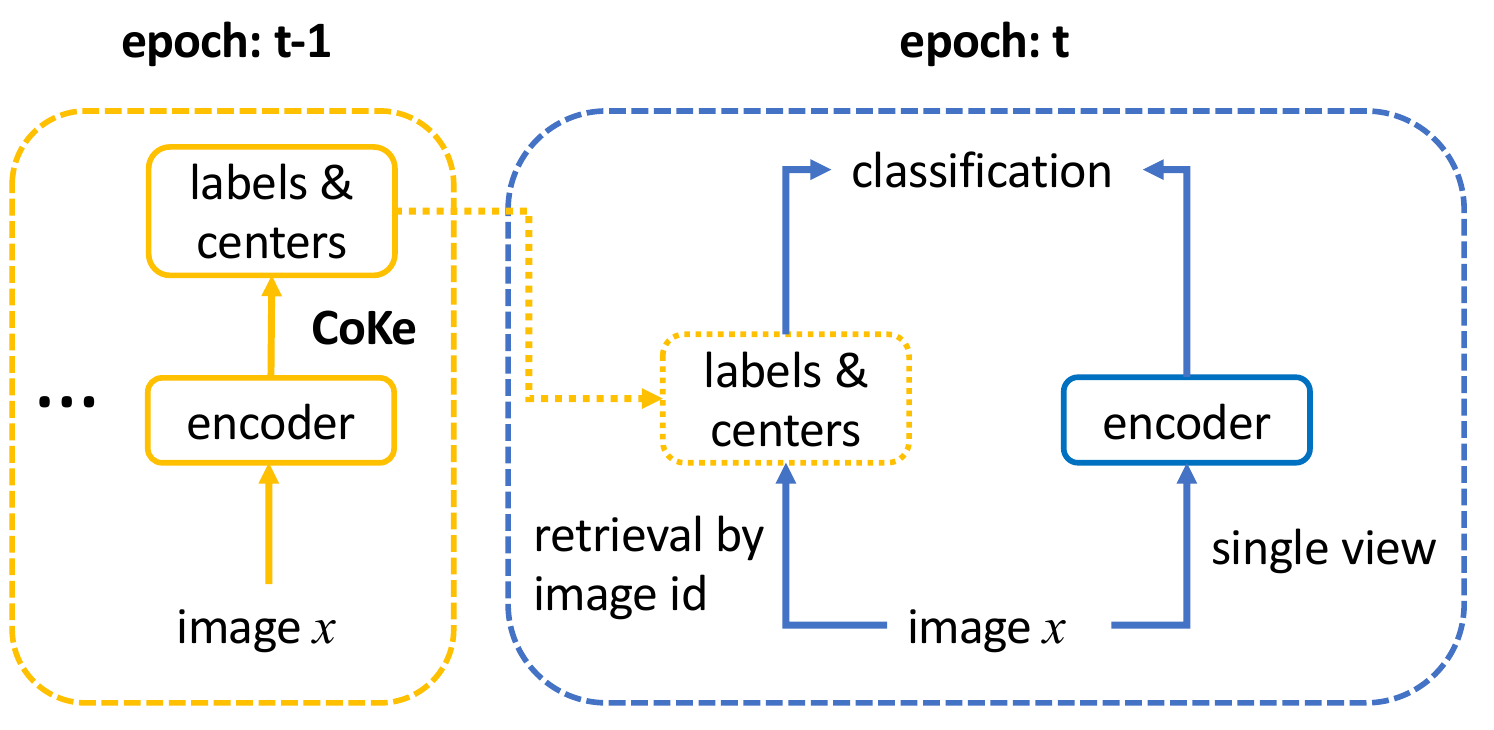}
\caption{Illustration of CoKe. When a mini-batch arrives, each instance will be assigned to a cluster with our online assignment method. Then, in epoch $t$, representations from the encoder network are optimized by discrimination using pseudo labels and cluster centers obtained from epoch $t-1$. The pseudo labels from epoch $t-1$ were stored to be retrieved in epoch $t$ using the unique id for each image.}\label{fig:illu}
\end{figure}

Besides instance discrimination, cluster discrimination is also an effective pretext task for unsupervised representation learning~\cite{AsanoRV20a,CaronBJD18,CaronMMGBJ20,abs-2005-04966,ZhuangZY19,ZhanX0OL20,abs-2104-14294}. Compared with instance discrimination that assigns a unique label to each instance, cluster discrimination partitions data into a pre-defined number of groups that is significantly less than the total number of instances. Therefore, the classification task after clustering becomes much more feasible for large-scale data. Furthermore, learning representations with clusters will push similar instances together, which may help explore potential semantic structures in data. Unfortunately, the clustering phase often needs to run multiple iterations over the entire data set, which has to be conducted in a batch mode to access representations of all instances~\cite{CaronBJD18}. Therefore, online clustering is adopted to improve the efficiency, while the collapsing problem (i.e., a dominating cluster that contains most of instances) becomes challenging for optimization. To mitigate the problem, ODC~\cite{ZhanX0OL20} has to memorize representations of all instances and decompose the dominating large cluster with a conventional batch mode clustering method. Instead, SwAV~\cite{CaronMMGBJ20} incorporates a balanced clustering method~\cite{AsanoRV20a} and obtains assignment with a batch mode solver for instances from only the last few mini-batches, which outperforms the vanilla online clustering in ODC~\cite{ZhanX0OL20} significantly. However, using only a small subset of data to generate pseudo labels can fail to capture the global distribution. Besides, balanced clustering constrains that each cluster has exactly the same number of instances, which can result in a suboptimal partition of data.

To take the benefits of cluster discrimination but mitigate the challenge, we, for the first time, investigate the objective of clustering-based representation learning from the perspective of distance metric learning~\cite{QianSSHTLJ19}. Our analysis shows that it indeed learns representations and relationships between instances simultaneously, while the coupled variables make the optimization challenging. By decoupling those variables appropriately, the problem can be solved in an alternating manner between two phases, that is, clustering and discrimination. When fixing representations, clustering is to discover relationship between instances. After that, the representations can be further refined by discrimination using labels from clustering. This finding explains the success of existing cluster discrimination methods. However, most existing methods have to conduct expensive clustering in a batch mode, while our analysis shows that an online method is feasible to optimize the objective. 

Based on the observation, we propose a novel pretext task with online \textbf{Co}nstrained \textbf{K}-m\textbf{e}ans (\textbf{CoKe}) for unsupervised representation learning. Concretely, in the clustering phase, we propose a novel online algorithm for constrained k-means that lower-bounds the size of each cluster. Different from balanced clustering, our strategy is more flexible to model inherent data structure. In addition, our theoretical analysis shows that the proposed online method can achieve a near-optimal assignment. In the discrimination phase, we adopt a standard normalized Softmax loss with labels and centers recorded from the last epoch to learn representations. By decoupling the clustering and discrimination phases, CoKe can learn representations with a \textit{single} view from each instance effectively and can be optimized with a small batch size. In addition, two variance reduction strategies are proposed to make the clustering robust for augmentations. Fig.~\ref{fig:illu} illustrates the framework of CoKe, which demonstrates a simple framework without additional components (e.g., momentum encoder~\cite{He0WXG20,GrillSATRBDPGAP20}, batch mode solver~\cite{CaronMMGBJ20,ZhanX0OL20}, etc.). Besides, only one label for each instance is kept in memory, which is an integer and the storage cost is negligible. 

Extensive experiments are conducted on both downstream tasks and clustering to demonstrate the proposal. With only a single view from each instance for training, CoKe already achieves a better performance than MoCo-v2~\cite{abs-2003-04297} that requires two views. By including additional views in optimization, CoKe demonstrates state-of-the-art performance on ImageNet and clustering.

\section{Related Work}\label{sec:related}
Various pretext tasks have been proposed for unsupervised representation learning. We briefly review instance discrimination and cluster discrimination that are closely related to our work, while other representative methods include BYOL~\cite{GrillSATRBDPGAP20}, SimSiam~\cite{ChenH21} and Barlow Twins~\cite{ZbontarJMLD21}.

\subsection{Instance Discrimination}
Instance discrimination is a straightforward pretext task for unsupervised representation learning, which tries to pull different augmentations from the same instance together but push them away from all other instances. Early work in this category optimizes the instance classification directly (i.e., each instance has one unique label), which implies an $N$-class classification problem, where $N$ is the total number of instances~\cite{DosovitskiyFSRB16}. Although promising results are obtained, this requires a large classification layer for deep learning. To improve the efficiency, the non-parametric contrastive loss is developed to mitigate the large-scale challenge~\cite{WuXYL18}. After that, many variants such as MoCo~\cite{He0WXG20,abs-2003-04297,abs-2104-02057} and SimCLR~\cite{ChenK0H20} are developed to approach or even outperform supervised pre-trained models on downstream tasks.

\subsection{Cluster Discrimination}
Instance discrimination focuses on individual instances and ignores the similarity between different instances. Therefore, clustering-based method is developed to capture the data structure better, which often consists of two phases: clustering and discrimination. DeepCluster~\cite{CaronBJD18} adopts a standard k-means for clustering, while SeLa~\cite{AsanoRV20a} proposes to solve an optimal transport problem for balanced assignment. After obtaining the pseudo labels, the representation will be learned by optimizing the corresponding classification problem. The bottleneck of these methods is that labels need to be assigned offline in a batch mode with representations of all instances to capture the global information. 

To reduce the cost of batch mode clustering, ODC~\cite{ZhanX0OL20} applies the standard online clustering to avoid the multiple iterations over the entire data set, while representations of all instances need to be kept in memory to address the collapsing problem. SwAV~\cite{CaronMMGBJ20} extends a batch mode optimal transport solver~\cite{AsanoRV20a} to do an online assignment to mitigate the collapsing problem. The assignment problem in SwAV is defined within a mini-batch of instances to save the storage for representations. To improve the effectiveness, the method stores representations from the last few mini-batches of instances to capture additional information. However, this is still a small subset compared to the whole data and thus the global information may not be exploited sufficiently. After that, DINO~\cite{abs-2104-14294} proposes to have the additional momentum encoder to stabilize the clustering without memorizing representations and can achieve the similar performance as SwAV with ResNet-50. 

Besides clustering-based pretext tasks, some work proposes to leverage nearest neighbors for each instance to capture semantic similarity between different instances~\cite{abs-2104-14548}. However, a large batch size and memory bank are required to capture the appropriate neighbors, which is expensive for optimization. In this work, we aim to improve the clustering phase with an online constrained k-means method, which gives better flexibility on cluster size and has a theoretical guarantee on the online assignment.

\section{Proposed Method}\label{sec:method}

\subsection{Objective for Clustering-Based Method}
We begin our analysis with the supervised representation learning. Given supervised label information, distance metric learning~\cite{WeinbergerS09} has been studied extensively to learn representations by optimizing triplet constraints including some efficient proxy-based variants~\cite{Attias17,QianSSHTLJ19,QianTLZJ18}. 
When there are $K$ classes in data, let $C = [\cc_1,\dots, \cc_K]\in\R^{d\times K}$ denote $K$ proxies with each corresponding to one class. 
The triplet constraint defined with proxies is $\forall \x_i,\cc_{k: k\not= y_i},\quad \|\x_i- \cc_k\|_2^2 - \|\x_i-\cc_{y_i}\|_2^2\geq \delta$, where $y_i$ is the label of $\x_i$. To maximize the margin, the optimization problem for supervised representation learning can be cast as
\begin{equation}\label{eq:supervised}
\min_{\x, C}  \sum_{i}\sum_{k:k\not=y_i} \|\x_i-\cc_{y_i}\|_2^2 -  \|\x_i-\cc_k\|_2^2
\end{equation}
which can be solved effectively with deep learning~\cite{QianSSHTLJ19}.

Without supervised label information, we assume that there are $K$ clusters in data. Besides proxies for each cluster, we have an additional variable $\mu$ such that $\mu_{i,k}=1$ assigning the $i$-th instance to the $k$-th cluster. We constrain the domain of $\mu$ as $\Delta=\{\mu|\forall i,\ \sum_{k}\mu_{i,k}=1,\forall i,k, \mu_{i,k}\in\{0,1\}\}$. It implies that each instance will only be assigned to a single cluster. The objective for the proxy-based unsupervised representation learning can be written as
\begin{scriptsize}
\begin{align}\label{eq:uloss}
\min_{\x, C, \mu\in\Delta}  \sum_{i}\Big((K-1)\sum_{k=1}^K \mu_{i,k}\|\x_i-\cc_k\|_2^2 -  \sum_{q=1}^K(1-\mu_{i,q})\|\x_i-\cc_q\|_2^2\Big)
\end{align}
\end{scriptsize}
The coupled variables in Eqn.~\ref{eq:uloss} make the optimization challenging. Hence, we can solve the problem in an alternating way. It should be noted that there are three groups of variables $\{\x, C, \mu\}$ and different decomposition can result in different algorithms.

We demonstrate a prevalent strategy that optimizes $\{\x\}$ and $\{\mu,C\}$ alternatively. When fixing assignment $\mu$ and centers $C$, the subproblem becomes
\begin{eqnarray*}
\min_{\x}  \sum_{i}\big(\sum_{q:q\not= \tilde{y}_i}^K \|\x_i-\cc_{\tilde{y}_i}\|_2^2 -  \|\x_i-\cc_q\|_2^2\big)
\end{eqnarray*}
where $\tilde{y}_i=\arg\max_k\mu_{i,k}$ is the pseudo label of the $i$-th instance. Given pseudo labels, it can be solved with a supervised method as for Eqn.~\ref{eq:supervised}, which is the \textit{discrimination phase} in representation learning.

When fixing representations $\x$, the subproblem can be simplified due to the empirical observation that the distribution of learned representations on the unit hypersphere has a mean that is close to zero~\cite{0001I20} as

\begin{eqnarray}\label{eq:kloss}
\min_{C,\mu\in \Delta}  \sum_{i} \sum_{k=1}^K \mu_{i,k}\|\x_i-\cc_k\|_2^2
\end{eqnarray}
It is a standard k-means clustering problem as the \textit{clustering phase} in representation learning. The analysis shows that decoupling clustering and discrimination~\cite{CaronBJD18, CaronMMGBJ20} is corresponding to an alternating solver for the objective in Eqn~\ref{eq:uloss}. In this work, we further decouple $\mu$ and $C$ in Eqn.~\ref{eq:kloss} for efficient online clustering.

\subsection{Online Constrained K-Means}
Since the clustering phase is more challenging, we address the problem in Eqn.~\ref{eq:kloss} first. As indicated in \cite{AsanoRV20a}, the original formulation may incur a trivial solution that most of instances go to the same cluster. To mitigate the problem, we adopt the constrained k-means~\cite{bradley2000constrained} instead, that is, controlling the minimal size of clusters to avoid collapsing.

Given a set of $N$ unlabeled data $\{\x_i\}$ and the number of clusters $K$, the objective for constrained k-means is
\begin{footnotesize}
\begin{align}\label{ckmeans}
\min_{C, \mu\in \Delta} \sum_{i=1,k=1}^{i=N,k=K}\mu_{i,k}\|\x_i - \cc_k\|_2^2 \quad s.t.\quad \forall k\quad \sum_{i=1}^N \mu_{i,k}\geq \gamma_k
\end{align}
\end{footnotesize}
where $\gamma_k$ is the lower-bound of cluster size for the $k$-the cluster. Consequently, the final objective for unsupervised representation learning becomes
\begin{scriptsize}
\begin{align}\label{eq:ucloss}
&\min_{\x, C, \mu\in\Delta}  \sum_{i}\Big((K-1)\sum_{k=1}^K \mu_{i,k}\|\x_i-\cc_k\|_2^2 -  \sum_{q=1}^K(1-\mu_{i,q})\|\x_i-\cc_q\|_2^2\Big)\nonumber\\
&s.t. \quad \forall k\quad \sum_{i=1}^N \mu_{i,k}\geq \gamma_k
\end{align}
\end{scriptsize}

The problem in Eqn.~\ref{ckmeans} can be solved in batch mode. However, neural networks are often optimized with stochastic gradient descent (SGD) that can access only a mini-batch of instances at each iteration. Therefore, we propose a novel online algorithm to handle the problem with a theoretical guarantee as follows.

\subsubsection{Online Assignment}
We consider the alternating solver for the problem in Eqn.~\ref{ckmeans}. When $C$ is fixed, the problem for updating $\mu$ can be simplified as an assignment problem
\begin{align}\label{eq:oa}
\max_{\mu\in\Delta'} \sum_i\sum_k s_{i,k} \mu_{i,k} \quad s.t.\quad \forall k\quad \sum_{i=1}^n \mu_{i,k}\geq \gamma_k
\end{align}
where the value of $\mu$ can be relaxed from the discrete space to the continuous space as $\mu_{i,k}\in[0,1]$ and $s_{i,k}$ is the similarity between the $i$-th instance and the $k$-th cluster. In this work we assume that $\x$ and $\cc$ have unit norm and thus $s_{i,k} = \x_i^\top \cc_k$.

Let $\mu^*$ denote the optimal solution for the problem in Eqn.~\ref{eq:oa}. The standard metric for online learning is
\begin{eqnarray*}
&&\R(\mu) = \sum_i\sum_k s_{i,k} \mu_{i,k}^* - \sum_i\sum_k s_{i,k} \mu_{i,k} \\ 
&&\V(\mu) = \max_k \{\gamma_k - \sum_i \mu_{i,k}\}
\end{eqnarray*}
where $\R(\mu)$ and $\V(\mu)$ denote regret and violation accumulated over $N$ instances, respectively. Since $\mu^*$ can be a solution with continuous values, the regret with $\mu^*$ is no less than that defined with a discrete assignment. Consequently, the performance to the optimal integer solution can be guaranteed if we can bound this regret well.

To solve the problem in Eqn.~\ref{eq:oa}, we first introduce a dual variable $\rho_k$ for each constraint $\sum_i \mu_{i,k} \geq \gamma_k$. To be consistent with the training scheme in deep learning, we assume that each instance arrives in a stochastic order. When the $i$-th instance arrives at the current iteration, the assignment can be obtained by solving the problem
\begin{eqnarray}\label{eq:as}
\max_{\mu_{i}\in\Delta'}  \sum_k s_{i,k} \mu_{i,k} + \sum_k\rho_k^{i-1} \mu_{i,k}
\end{eqnarray}
where $\{\rho_k^{i-1}\}$ are the dual variables from the last iteration and $\mu_i = [\mu_{i,1},\dots, \mu_{i,K}]$. The problem in Eqn.~\ref{eq:as} has a closed-form solution as
\begin{eqnarray}\label{eq:solution}
\mu_{i,k} = \left\{\begin{array}{cc}1&k=\arg\max_k s_{i,k}+\rho_k^{i-1}\\0&o.w.\end{array}\right.
\end{eqnarray}
Note that the domain for the assignment is a continuous space, but our solution implies an integer assignment. Besides, dual variables control the violation over the cluster size constraints. The method degrades to a greedy strategy without the dual variables.  

After assignment, dual variables will be updated as
\begin{eqnarray*}
\rho^i = \Pi_{\Delta_\tau}(\rho^{i-1} - \eta ( \mu_{i}-\frac{[\gamma_1,\dots,\gamma_K]}{N}))
\end{eqnarray*}
where $\Pi_{\Delta_\tau}$ projects the dual variables to the domain $\Delta_\tau = \{\rho| \forall k, \rho_k\geq0, \|\rho\|_1\leq \tau\}$. The performance of online assignment algorithm can be guaranteed in Theorem~\ref{thm:1}. Complete proofs can be found in the appendix.

\begin{thm}\label{thm:1}
If instances arrive in the stochastic order, by setting $\eta = \tau/\sqrt{2N}$, we have
\[E[\R(\mu)]\leq \OO(\sqrt{N}),\quad E[\V(\mu)]\leq \OO(\sqrt{N})\]
\end{thm}

\paragraph{Remark} Theorem~\ref{thm:1} indicates that compared with the optimal solution consisting of continuous assignment, the regret of our method with integer assignment can be well bounded. Besides, the violation is also bounded by $\OO(\sqrt{N})$ for the constraints accumulated over all instances. It illustrates that for each assignment, the gap to optimum can be bounded by $\OO(1/\sqrt{N})$ and our assignment method can achieve a near-optimal result even running online. Moreover, the theorem implies that the violation can be avoided by increasing $\gamma_k$ with a small factor.

For training with SGD, a mini-batch of instances rather than a single instance will arrive at each iteration. If the size of the mini-batch is $b$, we will assign pseudo labels for each instance with the closed-form solution in Eqn.~\ref{eq:solution}. The dual variables will be updated with the averaged gradient as
\begin{eqnarray}\label{eq:update}
\rho^i = \Pi_{\Delta_\tau}(\rho^{i-1} - \eta \frac{1}{b}\sum_{s=1}^b( \mu_{i}^s-\frac{[\gamma_1,\dots,\gamma_K]}{N}))
\end{eqnarray}

\subsubsection{Online Clustering}
With the proposed online assignment, we can update the assignment and centers for constrained k-means in an online manner. Concretely, for the $t$-th epoch, we first fix $C^{t-1}$ and assign pseudo labels for each mini-batch of instances. After training with an epoch of instances, the centers can be updated as
\begin{eqnarray}\label{eq:updateone}
\cc_k^t = \Pi_{\|\cc\|_2=1}(\frac{\sum_i^N\mu_{i,k}^t\x_i^t}{\sum_i^N \mu_{i,k}^t})
\end{eqnarray}
where $\mu^t$ is the assignment at the $t$-th epoch and $\x_i^t$ denotes a single view of the $i$-th instance at the $t$-th epoch.

Since our method does not memorize representations of instances, the variables in constrained k-means, especially centers, will only be updated once with an epoch of instances. However, k-means requires multiple iterations to converge as a batch mode method. Fortunately, clustering each epoch of data to optimum is not necessary for representation learning. According to the objective in Eqn.~\ref{eq:ucloss}, we can further decompose $\mu$ and $C$. When fixing $\x^t$ and $C^{t-1}$, the assignment can be updated by the proposed online assignment method. When fixing $\x^t$ and $\mu^t$, centers have a closed-form solution as in Eqn.~\ref{eq:updateone}. Therefore, a single step of updating is applicable for optimizing the target objective and the cost of clustering can be mitigated. Intuitively, representations are improved with more epochs of training while the clustering is gradually optimized simultaneously.

Furthermore, inspired by mini-batch k-means~\cite{Sculley10}, we can update the centers aggressively to accelerate the convergence of clustering process. Concretely, centers can be updated after each mini-batch as
\begin{eqnarray}\label{eq:updatemulti}
\cc_{k:m}^t = \Pi_{\|\cc\|_2=1}(\frac{\sum_i^m\mu_{i,k}^t\x_i^t}{\sum_i^m \mu_{i,k}^t})
\end{eqnarray}
where $m$ denotes the total number of received instances in the $t$-th epoch. After a sufficient training, we may switch to update centers only once in each epoch to reduce the variance from a mini-batch. Alg.~\ref{alg:ck} summarizes the proposed online clustering method.

\begin{algorithm}[t]
   \caption{Online \textbf{Co}nstrained \textbf{K}-M\textbf{e}ans (CoKe)}
   \label{alg:ck}
\begin{algorithmic}
   \STATE {\bfseries Input:} Data set $\{\x_i\}_{i=1}^N$, \#clusters $K$, \#epochs $T$, batch size $b$
   \STATE Randomly initialize $C^0$
   \FOR{$t=1$ {\bfseries to} $T$}
   \STATE Initialize $C_{0}^t = C^{t-1}$ and $m=0$
   \FOR{$r=1$ {\bfseries to} $N/b$}
   \STATE Obtain assignment $\mu^t$ as in (\ref{eq:solution})
   \STATE Update dual variables $\rho^{i}$ as in (\ref{eq:update})
   \STATE Update centers $C_{m+b}^t$ as in (\ref{eq:updatemulti})
   \STATE $m=m+b$
   \ENDFOR
   \ENDFOR
   \RETURN $\{\mu^T, C^T\}$
\end{algorithmic}
\end{algorithm}

\subsection{Discrimination}
With pseudo labels and centers obtained from the $(t-1)$-th epoch, we can learn representations by optimizing a standard normalized Softmax loss for instances at the $t$-th iteration as
\begin{eqnarray}\label{eq:loss}
\ell_{\mathrm{cls}}(\x_i^t) = -\log(\frac{\exp(\x_i^{t\top} \cc_{\tilde{y}_i^{t-1}}^{t-1}/\lambda)}{\sum_{k=1}^K \exp(\x_i^{t\top} \cc_{k}^{t-1}/\lambda)})
\end{eqnarray}
where $\tilde{y}_i^{t-1}$ is the pseudo label implied by $\mu^{t-1}$ and $\lambda$ is the temperature. Since $\mu_{i}$ is a one-hot vector, we can keep a single label for each instance in the memory, where the storage cost is negligible. $\x_i$ and $\cc_k$ have the unit norm. By decoupling clustering and discrimination, our method can optimize the objective in Eqn.~\ref{eq:ucloss} effectively in an alternating way. To initialize the pseudo labels and centers for representation learning, we scan one epoch of instances without training the model to obtain $\mu^0$ and $C^0$.

Finally, we show that our method converges. 
\begin{cor}
The proposed method will converge if keeping $\mu^{t-1}$ when $\mu^t$ provides no loss reduction.
\end{cor}
Although the theory requires to check the optimality of $\mu^t$, we empirically observe that CoKe works well with the vanilla implementation.

\subsection{Variance Reduction for Robust Clustering}

Variance from different views of each instance provides essential information for representation learning. However, it may perturb the clustering and make the optimization slow. Therefore, we propose two strategies to reduce the variance incurred to the assignment step.

\paragraph{Moving Average}
Ensemble is an effective way to reduce variance. Therefore, we propose to accumulate clustering results from the second stage. Concretely, for $t>T'$, assignment and centers will be updated as
\begin{eqnarray*}
&\hat{C}^{t} = (1-\frac{1}{t-T'})\hat{C}^{t-1}+\frac{1}{t-T'}C^{t}; \nonumber\\
&\hat{\y}^{t} = (1-\frac{1}{t-T'})\hat{\y}^{t-1}+\frac{1}{t-T'}\tilde{\y}^{t}
\end{eqnarray*}
where $C^{t}$ and $\tilde{y}^t$ are obtained at the $t$-th epoch and $\tilde{\y}^t$ denotes the one-hot vector of $\tilde{y}^t$. The formulation averages the clustering results from the last $T-T'$ epochs to reduce the variance from augmentations. Unlike $\tilde{\y}_i$, $\hat{\y}_i$ is not a one-hot vector due to ensemble and can contain multiple non-zero terms. We adopt the loss defined with soft labels as
\begin{eqnarray*}
\ell_{\mathrm{cls}}^{\mathrm{soft}}(\x_i^{t}) = -\sum_{k}\hat{\y}_{i,k}^{t-1}\log(\frac{\exp(\x_i^{t\top} \tilde{\cc}_{k}^{t-1}/\lambda)}{\sum_{j=1}^K \exp(\x_i^{t\top} \tilde{\cc}_{j}^{t-1}/\lambda)})
\end{eqnarray*}

\paragraph{Two Views}
Learning representations with two views from the same image is prevalent in contrastive learning. Our proposed method can be considered as leveraging two views from different epochs and thus a single view is sufficient for each epoch. Nevertheless, CoKe can be further improved by accessing two views at each iteration. 

Given two views of an image, the constraint for assignment is that both views share the same label. Therefore, the assignment problem in Eqn.~\ref{eq:as} becomes
\begin{eqnarray*}
\max_{\mu_{i}\in\Delta'} \frac{1}{2}\sum_k\mu_{i,k}\sum_{j=1}^{2}  s_{i,k}^j  + \sum_k\rho_k^{i-1} \mu_{i,k}
\end{eqnarray*}
where $s_{i,k}^j$ denotes the similarity between the $j$-th view of the $i$-th instance and the $k$-th center. Hence, it is equivalent to obtaining a label for the mean vector averaged over two views as
\begin{eqnarray}\label{eq:twomu}
\mu_{i,k} = \left\{\begin{array}{cc}1&k=\arg\max_k \frac{1}{2}\sum_{j=1}^2 s_{i,k}^j+\rho_k^{i-1}\\0&o.w.\end{array}\right.
\end{eqnarray}
Then, the loss in Eqn.~\ref{eq:loss} will be averaged over two views. Compared with the single view, multiple views can reduce the variance from different augmentations and make the assignment more stable. 

Besides variance reduction for the one-hot assignment, the other advantage with the additional view is that it can provide a reference label distribution for the other view. Let $p_{i:j}$ denote the predicted probability over labels
\[p_{i:j,q}^{t-1} = \frac{\exp(\x_i^{j\top} \cc_q^{t-1}/\lambda)}{\sum_{k=1}^K \exp(\x_i^{j\top} \cc_{k}^{t-1}/\lambda)}\]
We can obtain the soft label for view $1$ with the reference from view $2$ as \[\hat{\y}_{i:1}^{t} = \alpha\tilde{\y}_i^{t-1} + (1-\alpha)\p_{i:2}^{t-1}\]
Then, the cross entropy loss for view $1$ can be optimized with $\hat{\y}_{i:1}^{t}$ instead. Alg.~\ref{alg:code} summarizes the pseudo-code of CoKe with two views, which can be extended to multiple views easily.

\begin{algorithm}[t]
\caption{Pseudo-code of CoKe with Two Views.}
\label{alg:code}
\definecolor{codeblue}{rgb}{0.25,0.5,0.5}
\lstset{
  backgroundcolor=\color{white},
  basicstyle=\fontsize{7.2pt}{7.2pt}\ttfamily\selectfont,
  columns=fullflexible,
  breaklines=true,
  captionpos=b,
  commentstyle=\fontsize{7.2pt}{7.2pt}\color{codeblue},
  keywordstyle=\fontsize{7.2pt}{7.2pt},
}
\begin{lstlisting}[language=python]
# f: encoder network for input images
# u: pseudo one-hot labels (Nx1)
# C: cluster centers
# rho: dual variable for constraints (Kx1)
# gamma: lower-bound of cluster size
# lambda: temperature
# alpha: ratio between labels

for z in loader:  # load a minibatch with b samples
    z_1, z_2 = aug(z), aug(z) # two random views from z
    x_1, x_2 = f(z_1), f(z_2) # encoder representations
    s_1, s_2 = x_1C, x_2C # logits over centers
    y = u(z_id) # retrieve label from last epoch
    # compute reference distribution for each view
    p_1 = softmax(s_1/lambda)
    p_2 = softmax(s_2/lambda)
    # obtain soft label for discrimination
    y_1 = alpha*y + (1-alpha)*p_2 
    y_2 = alpha*y + (1-alpha)*p_1
    # loss over two views
    loss = 0.5*(-y_1*log(p_1) -y_2*log(p_2)) 
    loss.backward() # update encoder
    # update clustering
    x_mean = 0.5*(x_1+x_2) # mean vector of two views
    u(z_id) = update(x_mean, C, rho) # as in Eqn. 13
    C = update(C, x_mean, u(z_id)) # as in Eqn. 11
    rho = update(rho, gamma, u(z_id)) # as in Eqn. 14
\end{lstlisting}
\end{algorithm}

\section{Experiments}\label{sec:exp}
We conduct experiments of unsupervised representation learning on ImageNet~\cite{RussakovskyDSKS15} to evaluate the proposed method. For fair comparison, we follow settings in benchmark methods~\cite{CaronMMGBJ20,ChenK0H20,abs-2003-04297}. More details can be found in the appendix. 

For the parameters in CoKe, we set the learning rate as $1.6$ and temperature $\lambda=0.1$. Besides the learning rate for model, CoKe contains another learning rate $\eta$ for updating dual variables as in Eqn.~\ref{eq:update}. We empirically observe that it is insensitive and set $\eta=20$. Finally, the batch size is $1,024$ such that all experiments of CoKe except the one with multi-crop can be implemented on a standard server with $8$ GPUs and $16$G memory on each GPU.

An important parameter in CoKe is the minimal cluster size. To reduce the number of parameters, we assign the same constraint for different clusters as $\gamma_1=\dots=\gamma_K=\gamma$. Considering that $\gamma = N/K$ denotes the balanced clustering, we introduce a parameter $\gamma'$ as $\gamma = \gamma' N/K$ and tune $\gamma'$ in lieu of $\gamma$ for better illustration. In the experiments, we observe that the maximal value of dual variables is well bounded, so we simplify the updating criterion for dual variables as 
\begin{eqnarray}\label{eq:finalrho}
\rho_k^i = \max\{0,\rho_k^{i-1} - \eta \frac{1}{b}\sum_{s=1}^b( \mu_{i,k}^s-\frac{\gamma'}{K})\}
\end{eqnarray}

\subsection{Ablation Study}
First, we empirically study the effect of each component in CoKe. All experiments in this subsection train $200$ epochs and each instance has a single view of augmentation at each iteration. After obtaining the model, the learned representations are evaluated by learning a linear classifier on ImageNet. The training protocol for linear classifier follows that in MoCo~\cite{He0WXG20} except that we change the weight decay to $10^{-6}$ and learning rate to $1$ for our pre-trained model. 

\subsubsection{Balanced vs. Constrained Clustering}
In the previous work~\cite{AsanoRV20a,CaronMMGBJ20}, balanced clustering that constrains each cluster to have the same number of instances demonstrates a good performance for representation learning. Constrained clustering that lower-bounds the size of each cluster is a more generic setting, but has been less investigated. With the proposed method, we compare constrained clustering to balanced clustering in Table~\ref{ta:ratio}. 

\begin{table}[!ht]
\centering
\begin{tabular}{|l|l|l|l|l|}\hline
Ratio: $\gamma'$ &Acc\%&\#Cons&\#Min&\#Max\\\hline
1 &63.1 &427&403&445\\\hline
0.8&63.8&342&338&1,301\\\hline
0.6&64.3&256&254&1,404\\\hline
0.4&64.5&171&168&2,371\\\hline
0&41.3&0&0&449$k$\\\hline
\end{tabular}
\caption{Comparison of different ratios $\gamma'$ in CoKe. The performance is evaluated by linear classification with learned representations on ImageNet as in MoCo~\cite{He0WXG20}.}\label{ta:ratio}
\end{table}

We fix the number of centers as $K=3,000$ while varying $\gamma'$ to evaluate the effect of cluster size constraint. When $\gamma'=1$, each cluster has to contain $N/K$ instances that becomes the balanced clustering. We let ``\#Cons'', ``\#Min'', ``\#Max'' denote the constrained cluster size, the actual size of the smallest cluster and that of the largest cluster from the last epoch of CoKe, respectively. As illustrated in Table~\ref{ta:ratio}, the balanced clustering can achieve $63.1\%$ accuracy when training with a single view. It confirms that balanced clustering is effective for learning representations. If decreasing the ratio, each cluster can have a different number of instances that is more flexible to capture the inherent data structure. For example, when $\gamma'=0.8$, the minimum size of clusters is reduced from $403$ to $338$ while the largest cluster has more than double of instances in balanced clustering. Meanwhile, the imbalanced partition helps to improve the accuracy by $0.7\%$. With an even smaller ratio of $0.4$, our method surpasses the balanced clustering with a significant margin of $1.4\%$ and it demonstrates that constrained clustering is more appropriate for unsupervised representation learning. The performance will degrade when $\gamma'=0$ since it may incur the collapsing problem without a sufficient number of instances in each cluster. We will fix $\gamma'=0.4$ in the following experiments.  

Besides the accuracy on linear classification, we further investigate the violation of constraints in Table~\ref{ta:ratio}. For balanced clustering, each cluster has the same number of instances which is a strong constraint. Compared to the constraint, the violation of our online assignment is only $5\%$ when $\gamma'=1$. If $\gamma'$ is less than $1$, the constraint is relaxed and the violation can be reduced to less than $1\%$, which illustrates the effectiveness of our method. Compared with the online assignment strategy that only optimizes the constraints over a small subset of data in SwAV~\cite{CaronMMGBJ20}, we optimize the assignment globally and can explore the distribution of data sufficiently. Interestingly, we find that there is no dominating cluster even when $\gamma'=0.4$. In that scenario, the largest cluster only contains $2,371$ instances. It illustrates that clustering is effective to learn an appropriate partition for unlabeled data. If $\gamma'=0$, more than $449,000$ instances will be assigned to the same cluster, which confirms the importance of cluster size constraint to mitigate the collapsing problem.

\subsubsection{Coupled Clustering and Discrimination}
Then, we study the effect of coupling clustering and discrimination. In CoKe, we decouple clustering and discrimination by collecting clustering results from the last epoch to discriminate data from the current epoch. Table~\ref{ta:lc} compares the performance with different labels and centers where $\{C^{t-1}, \tilde{y}^{t-1}\}$ and $\{C^{t}, \tilde{y}^{t}\}$ are from the last epoch and the current epoch, respectively.

\begin{table}[!ht]
\centering
\scriptsize
\begin{tabular}{|l|c|c|c|c|}\hline
Settings&$\{C^{t-1}, \tilde{y}^{t-1}\}$&$\{C^{t-1}, \tilde{y}^{t}\}$&$\{C^{t}, \tilde{y}^{t-1}\}$&$\{C^{t}, \tilde{y}^{t}\}$\\\hline
Acc\% &64.5&0.4& 51.2&0.1\\\hline
\end{tabular}
\caption{Comparison of labels and centers from different epochs.}\label{ta:lc}
\end{table}

First, we can observe that with labels and centers from the last epoch, CoKe demonstrates the best performance. It verifies that CoKe solves the problem in Eqn.~\ref{eq:ucloss} effectively in an alternating way. Second, with current centers $C^t$, the performance decreases more than $10\%$, which shows the importance of keeping centers from the last epoch in CoKe. Finally, the other two variants with $\tilde{y}^{t}$ fail to learn meaningful representations. It is consistent with our analysis for the objective in Eqn.~\ref{eq:ucloss}. Note that $\mu$ is the additional variables introduced by unsupervised learning and decoupling $\x$ and $\mu$ is crucial for clustering-based representation learning.

\subsubsection{Number of Clusters}
The number of clusters is a key parameter in k-means. When $K$ is small, the relationship between similar instances may not be exploited sufficiently. However, additional noise can be introduced with a large $K$. Instance classification can be considered as a special case when $K=N$. Table~\ref{ta:k} summarizes the performance with different $K$'s. We observe that CoKe with $1,000$ clusters is about $1\%$ worse than that with $K=3,000$. It is because that a small $K$ is hard to capture all informative patterns due to a coarse granularity. 

\begin{table}[!ht]
\centering
\begin{tabular}{|l|l|l|l|l|}\hline
$K$&Acc\%&\#Cons&\#Min&\#Max\\\hline
1,000&63.4&512&512&4,639\\\hline
3,000&64.5&171&168&2,371\\\hline
5,000&64.3&102&98&1,982\\\hline
\end{tabular}
\caption{Comparison of number of clusters $K$ in k-means.}\label{ta:k}
\end{table}

However, obtaining an appropriate $K$ for clustering is a challenging problem in k-means. Moreover, clustering can provide different results even with the same representations, which is researched in multi-clustering~\cite{HuQPJZ17,hu2018subspace}. This phenomenon is due to the fact that objects can be similar in different ways (e.g., color, shape, etc.). Multi-clustering has been explored in previous representation learning work~\cite{AsanoRV20a,abs-2005-04966} and we also apply it to learn representations with a multi-task framework. Each task is defined as a constrained k-means problem with a different $K$, while the final loss will be averaged over multiple tasks. This strategy mitigates the parameter setting problem in k-means by handling multiple k-means problems with diverse parameters simultaneously.

\begin{table}[!ht]
\centering
\begin{tabular}{|l|l|l|l|l|l|l|}\hline
$K$($\times 1,000$)&3&2+3&3+4&3+3+3&3+4+5\\\hline
Acc\% &64.5&65.0&65.2&65.2&65.3\\\hline
\end{tabular}
\caption{Multi-clustering with different $K$ combinations.}\label{ta:multik}
\end{table}

Table~\ref{ta:multik} shows the results of learned representations with multi-clustering. When including a task with $K=2,000$, the accuracy is improved from $64.5\%$ to $65.0\%$. With a more fine-grained task of $K=4,000$, the performance of learned representations is even better and achieves $65.2\%$ in accuracy. Then, we evaluate the triple k-means task with two different settings, that is, same $K$ and different $K$'s. It can be observed that different $K$'s can further improve the performance. We will adopt the strategy in rest experiments for the explicit multi-clustering. More ablation study can be found in the appendix.

\subsection{Comparison with State-of-the-Art on ImageNet}
In this subsection, we compare our proposal with state-of-the-art methods by learning a linear classifier on learned representations for ImageNet. All methods have ResNet-50 as the backbone. The results of methods with similar configurations (e.g., $2$-layer MLP, $128$-dimensional representations, etc.) are summarized in Table~\ref{ta:sota1}. 

\begin{table}[!ht]
\centering
\begin{tabular}{|l|l|l|l|l|}\hline
Methods&\#View&\#Epoch&\#Dim&Acc\%\\\hline
SimCLR&2&1,000&128& 69.3\\
MoCo-v2&2&800&128&71.1\\
DeepCluster-v2&2&400&128&70.2\\
SwAV&2&400&128&70.1\\\hline
CoKe&1&800&128&\textbf{71.4}\\\hline
\end{tabular}
\caption{Comparison with methods that have the similar configuration on ImageNet by linear classification.}\label{ta:sota1}
\end{table}

Explicitly, baseline methods have to learn representations with two views of augmentations from an individual instance at each iteration for the desired performance. On the contrary, CoKe can work with a single view using online optimization. It can be observed that the accuracy of representations learned by CoKe with $800$ epochs can achieve $71.4\%$, which performs slightly better than MoCo-v2 but with only a half number of views for optimization. It illustrates that leveraging relations between instances can learn more informative patterns than instance discrimination. Second, compared to the clustering-based methods, CoKe outperforms SwAV and DeepCluster by $1\%$ when training with the same number of views. This further demonstrates the effectiveness of CoKe. Finally, we compare the running time for training one epoch of data in Table~\ref{ta:time}. With a single view for optimization, the learning efficiency can be significantly improved as in CoKe. 

\begin{table}[!ht]
\centering
\begin{tabular}{|l|l|l|l|}\hline
MoCo-v2&SwAV&CoKe&CoKe*\\\hline
18.3&20.8& 11.1 &8.4 \\\hline
\end{tabular}
\caption{Comparison of running time (mins) for training one epoch of data on ImageNet. All methods are evaluated on the same server. CoKe* applies automatic mixed precision training provided by PyTorch.}\label{ta:time}
\end{table}

Then, we apply more sophisticated settings proposed by recent methods~\cite{GrillSATRBDPGAP20, ChenKSNH20} for CoKe and compare with methods using different settings. Concretely, we include $3$-layer MLP, an additional 2-layer prediction head and $1,000$ epochs for training. More details can be found in the appendix. Table~\ref{ta:sota2} summarizes the comparison.

\begin{table}[!ht]
\centering
\footnotesize
\begin{tabular}{|l|l|l|l|l|l|l|}\hline
Methods&\#V&Bs&\#D&ME&MB&Acc\%\\\hline
SimSiam~\cite{ChenH21}&2&256&2,048&\checkmark&&71.3\\
SwAV~\cite{ChenH21}&2&4,096&128&&&71.8\\
MoCo-v2+~\cite{ChenH21}&2&256&128&\checkmark&\checkmark&72.2\\
Barlow Twins~\cite{ZbontarJMLD21}&2&2,048&8,192&&&73.2\\
MoCo-v3~\cite{abs-2104-02057}&2&4,096&256&\checkmark&&73.8\\
BYOL~\cite{GrillSATRBDPGAP20}&2&4,096&256&\checkmark&&74.3\\
NNCLR~\cite{abs-2104-14548}&2&1,024&256&\checkmark&\checkmark&72.9\\
NNCLR~\cite{abs-2104-14548}&2&4,096&256&\checkmark&\checkmark&75.4\\
\hline
DeepCluster-v2~\cite{CaronMMGBJ20}&8&4,096&128&&&75.2\\
SwAV~\cite{CaronMMGBJ20}&8&4,096&128&&&75.3\\
DINO~\cite{abs-2104-14294}&8&4,096&256&\checkmark&&75.3\\
NNCLR~\cite{abs-2104-14548}&8&4,096&256&\checkmark&\checkmark&75.6\\
\hline
CoKe&1&1,024&128&&&72.5\\
CoKe&2&1,024&128&&&74.9\\
CoKe&8&1,024&128&&&\textbf{76.4}\\\hline
\end{tabular}
\caption{Comparison with state-of-the-art methods on ImageNet by linear classification. ME and MB denote momentum encoder and memory bank, respectively.}\label{ta:sota2}
\end{table}

First, we can observe that CoKe with single view performs slightly better than MoCo-v2 again and it demonstrates that optimizing with single view is able to obtain an applicable pre-trained model. Second, by equipping with two views, CoKe can achieve $74.9\%$ accuracy on ImageNet, which is a competitive result but with much lighter computational cost. Furthermore, the superior performance of NNCLR and CoKe shows that capturing relations between instances can learn better representations. However, NNCLR has to obtain appropriate nearest neighbors with a large memory bank and is sensitive to the batch size. On the contrary, CoKe learns relationship by online clustering, which is feasible for small batch size and leads to a simple framework without memory bank and momentum encoder. Finally, with the standard multi-crop trick, CoKe can achieve $76.4\%$ accuracy on ImageNet that is close to the supervised counterpart, i.e., $76.5\%$. In summary, CoKe is more resource friendly (e.g., a standard server with $8$ GPUs is sufficient) with superb performance.

\subsection{Comparison on Downstream Tasks}
Besides linear classification on ImageNet, we evaluate CoKe on various downstream tasks in Table~\ref{ta:ds}. Methods with public available pre-trained models are included for comparison. For a fair comparison, we search parameters for all baselines with the codebase from MoCo. Evidently, CoKe provides a better performance than the strong baselines with multi-crop training, which confirms the effectiveness of our method. Detailed empirical settings and additional experiments on clustering are in the appendix.

\begin{table}[!ht]
\centering
\small
\begin{tabular}{|l|l|l|l|l|l|}\hline
&VOC&\multicolumn{2}{c|}{COCO}&C10&C100\\\hline
Methods&Ap$_{50}$&Ap$^{bb}$&Ap$^{mk}$&Acc\%&Acc\%\\\hline
Supervised&81.3&38.9&35.4 &97.3&86.6\\\hline
MoCo-v2&\underline{83.0}&39.6&35.9&97.9&86.1\\\hline
Barlow Twins&81.5&40.1&36.9&98.0&87.4 \\\hline
BYOL& 82.9&\underline{40.5}&36.9&\underline{98.1}&\underline{87.9}  \\\hline
SwAV$^*$&82.1&40.4&\underline{37.1}&97.7&87.5 \\\hline
DINO$^*$&82.0&40.2&36.8&97.7& 87.6\\\hline
CoKe&\underline{83.2}&\underline{40.9}&\underline{37.2}&\underline{98.2}&\underline{88.2}\\\hline
\end{tabular}
\caption{Comparison on downstream tasks. $^*$ denotes the multi-crop training trick. Top-2 best models are underlined.}\label{ta:ds}
\end{table}

\section{Conclusion}\label{sec:conclude}
In this work, we propose a novel learning objective for cluster discrimination pretext task. An online constrained k-means method with theoretical guarantee is developed to obtain pseudo labels, which is more appropriate for stochastic training in representation learning. The empirical study shows that CoKe can learn effective representations with less computational cost by leveraging the aggregation information between similar instances. Recently, Transformer~\cite{DosovitskiyB0WZ21} shows the superior performance, evaluating CoKe on the new architecture can be our future work.

{\small
\bibliographystyle{ieee_fullname}
\bibliography{coke}

\begin{thebibliography}{10}\itemsep=-1pt

\bibitem{AsanoRV20a}
Yuki~Markus Asano, Christian Rupprecht, and Andrea Vedaldi.
\newblock Self-labelling via simultaneous clustering and representation
  learning.
\newblock In {\em ICLR}, 2020.

\bibitem{bradley2000constrained}
Paul~S Bradley, Kristin~P Bennett, and Ayhan Demiriz.
\newblock Constrained k-means clustering.
\newblock {\em Microsoft Research, Redmond}, 20(0):0, 2000.

\bibitem{CaronBJD18}
Mathilde Caron, Piotr Bojanowski, Armand Joulin, and Matthijs Douze.
\newblock Deep clustering for unsupervised learning of visual features.
\newblock In {\em ECCV}, 2018.

\bibitem{CaronMMGBJ20}
Mathilde Caron, Ishan Misra, Julien Mairal, Priya Goyal, Piotr Bojanowski, and
  Armand Joulin.
\newblock Unsupervised learning of visual features by contrasting cluster
  assignments.
\newblock In {\em NeurIPS}, 2020.

\bibitem{abs-2104-14294}
Mathilde Caron, Hugo Touvron, Ishan Misra, Herv{\'{e}} J{\'{e}}gou, Julien
  Mairal, Piotr Bojanowski, and Armand Joulin.
\newblock Emerging properties in self-supervised vision transformers.
\newblock {\em CoRR}, abs/2104.14294, 2021.

\bibitem{ChenK0H20}
Ting Chen, Simon Kornblith, Mohammad Norouzi, and Geoffrey~E. Hinton.
\newblock A simple framework for contrastive learning of visual
  representations.
\newblock In {\em ICML}, volume 119, pages 1597--1607, 2020.

\bibitem{ChenKSNH20}
Ting Chen, Simon Kornblith, Kevin Swersky, Mohammad Norouzi, and Geoffrey~E.
  Hinton.
\newblock Big self-supervised models are strong semi-supervised learners.
\newblock In Hugo Larochelle, Marc'Aurelio Ranzato, Raia Hadsell,
  Maria{-}Florina Balcan, and Hsuan{-}Tien Lin, editors, {\em NeurIPS}, 2020.

\bibitem{abs-2003-04297}
Xinlei Chen, Haoqi Fan, Ross~B. Girshick, and Kaiming He.
\newblock Improved baselines with momentum contrastive learning.
\newblock {\em CoRR}, abs/2003.04297, 2020.

\bibitem{ChenH21}
Xinlei Chen and Kaiming He.
\newblock Exploring simple siamese representation learning.
\newblock In {\em CVPR}, pages 15750--15758. Computer Vision Foundation /
  {IEEE}, 2021.

\bibitem{abs-2104-02057}
Xinlei Chen, Saining Xie, and Kaiming He.
\newblock An empirical study of training self-supervised vision transformers.
\newblock {\em CoRR}, abs/2104.02057, 2021.

\bibitem{DosovitskiyB0WZ21}
Alexey Dosovitskiy, Lucas Beyer, Alexander Kolesnikov, Dirk Weissenborn,
  Xiaohua Zhai, Thomas Unterthiner, Mostafa Dehghani, Matthias Minderer, Georg
  Heigold, Sylvain Gelly, Jakob Uszkoreit, and Neil Houlsby.
\newblock An image is worth 16x16 words: Transformers for image recognition at
  scale.
\newblock In {\em ICLR}. OpenReview.net, 2021.

\bibitem{DosovitskiyFSRB16}
Alexey Dosovitskiy, Philipp Fischer, Jost~Tobias Springenberg, Martin~A.
  Riedmiller, and Thomas Brox.
\newblock Discriminative unsupervised feature learning with exemplar
  convolutional neural networks.
\newblock {\em {IEEE} Trans. Pattern Anal. Mach. Intell.}, 38(9):1734--1747,
  2016.

\bibitem{abs-2104-14548}
Debidatta Dwibedi, Yusuf Aytar, Jonathan Tompson, Pierre Sermanet, and Andrew
  Zisserman.
\newblock With a little help from my friends: Nearest-neighbor contrastive
  learning of visual representations.
\newblock {\em CoRR}, abs/2104.14548, 2021.

\bibitem{EveringhamGWWZ10}
Mark Everingham, Luc~Van Gool, Christopher K.~I. Williams, John~M. Winn, and
  Andrew Zisserman.
\newblock The pascal visual object classes {(VOC)} challenge.
\newblock {\em Int. J. Comput. Vis.}, 88(2):303--338, 2010.

\bibitem{GansbekeVGPG20}
Wouter~Van Gansbeke, Simon Vandenhende, Stamatios Georgoulis, Marc Proesmans,
  and Luc~Van Gool.
\newblock {SCAN:} learning to classify images without labels.
\newblock In Andrea Vedaldi, Horst Bischof, Thomas Brox, and Jan{-}Michael
  Frahm, editors, {\em ECCV}, volume 12355 of {\em Lecture Notes in Computer
  Science}, pages 268--285. Springer, 2020.

\bibitem{GrillSATRBDPGAP20}
Jean{-}Bastien Grill, Florian Strub, Florent Altch{\'{e}}, Corentin Tallec,
  Pierre~H. Richemond, Elena Buchatskaya, Carl Doersch, Bernardo~{\'{A}}vila
  Pires, Zhaohan Guo, Mohammad~Gheshlaghi Azar, Bilal Piot, Koray Kavukcuoglu,
  R{\'{e}}mi Munos, and Michal Valko.
\newblock Bootstrap your own latent - {A} new approach to self-supervised
  learning.
\newblock In {\em NeurIPS}, 2020.

\bibitem{He0WXG20}
Kaiming He, Haoqi Fan, Yuxin Wu, Saining Xie, and Ross~B. Girshick.
\newblock Momentum contrast for unsupervised visual representation learning.
\newblock In {\em CVPR}, pages 9726--9735, 2020.

\bibitem{HeGDG17}
Kaiming He, Georgia Gkioxari, Piotr Doll{\'{a}}r, and Ross~B. Girshick.
\newblock Mask {R-CNN}.
\newblock In {\em ICCV}, pages 2980--2988, 2017.

\bibitem{HeZRS16}
Kaiming He, Xiangyu Zhang, Shaoqing Ren, and Jian Sun.
\newblock Deep residual learning for image recognition.
\newblock In {\em CVPR}, pages 770--778, 2016.

\bibitem{hu2018subspace}
Juhua Hu and Jian Pei.
\newblock Subspace multi-clustering: a review.
\newblock {\em Knowledge and information systems}, 56(2):257--284, 2018.

\bibitem{HuQPJZ17}
Juhua Hu, Qi Qian, Jian Pei, Rong Jin, and Shenghuo Zhu.
\newblock Finding multiple stable clusterings.
\newblock {\em Knowl. Inf. Syst.}, 51(3):991--1021, 2017.

\bibitem{IoffeS15}
Sergey Ioffe and Christian Szegedy.
\newblock Batch normalization: Accelerating deep network training by reducing
  internal covariate shift.
\newblock In {\em ICML}, volume~37, pages 448--456, 2015.

\bibitem{JiVH19}
Xu Ji, Andrea Vedaldi, and Jo{\~{a}}o~F. Henriques.
\newblock Invariant information clustering for unsupervised image
  classification and segmentation.
\newblock In {\em ICCV}, pages 9864--9873. {IEEE}, 2019.

\bibitem{krizhevsky2009learning}
Alex Krizhevsky and Geoffrey Hinton.
\newblock Learning multiple layers of features from tiny images.
\newblock 2009.

\bibitem{abs-2005-04966}
Junnan Li, Pan Zhou, Caiming Xiong, Richard Socher, and Steven C.~H. Hoi.
\newblock Prototypical contrastive learning of unsupervised representations.
\newblock {\em CoRR}, abs/2005.04966, 2020.

\bibitem{LinMBHPRDZ14}
Tsung{-}Yi Lin, Michael Maire, Serge~J. Belongie, James Hays, Pietro Perona,
  Deva Ramanan, Piotr Doll{\'{a}}r, and C.~Lawrence Zitnick.
\newblock Microsoft {COCO:} common objects in context.
\newblock In {\em ECCV}, volume 8693, pages 740--755, 2014.

\bibitem{Attias17}
Yair Movshovitz{-}Attias, Alexander Toshev, Thomas~K. Leung, Sergey Ioffe, and
  Saurabh Singh.
\newblock No fuss distance metric learning using proxies.
\newblock In {\em ICCV}, pages 360--368, 2017.

\bibitem{NorooziF16}
Mehdi Noroozi and Paolo Favaro.
\newblock Unsupervised learning of visual representations by solving jigsaw
  puzzles.
\newblock In Bastian Leibe, Jiri Matas, Nicu Sebe, and Max Welling, editors,
  {\em ECCV}, volume 9910, pages 69--84, 2016.

\bibitem{PathakKDDE16}
Deepak Pathak, Philipp Kr{\"{a}}henb{\"{u}}hl, Jeff Donahue, Trevor Darrell,
  and Alexei~A. Efros.
\newblock Context encoders: Feature learning by inpainting.
\newblock In {\em CVPR}, pages 2536--2544, 2016.

\bibitem{QianSSHTLJ19}
Qi Qian, Lei Shang, Baigui Sun, Juhua Hu, Hao Li, and Rong Jin.
\newblock Softtriple loss: Deep metric learning without triplet sampling.
\newblock In {\em ICCV}, pages 6449--6457, 2019.

\bibitem{QianTLZJ18}
Qi Qian, Jiasheng Tang, Hao Li, Shenghuo Zhu, and Rong Jin.
\newblock Large-scale distance metric learning with uncertainty.
\newblock In {\em CVPR}, pages 8542--8550, 2018.

\bibitem{RenHG017}
Shaoqing Ren, Kaiming He, Ross~B. Girshick, and Jian Sun.
\newblock Faster {R-CNN:} towards real-time object detection with region
  proposal networks.
\newblock {\em {IEEE} Trans. Pattern Anal. Mach. Intell.}, 39(6):1137--1149,
  2017.

\bibitem{RussakovskyDSKS15}
Olga Russakovsky, Jia Deng, Hao Su, Jonathan Krause, Sanjeev Satheesh, Sean Ma,
  Zhiheng Huang, Andrej Karpathy, Aditya Khosla, Michael~S. Bernstein,
  Alexander~C. Berg, and Fei{-}Fei Li.
\newblock Imagenet large scale visual recognition challenge.
\newblock {\em Int. J. Comput. Vis.}, 115(3):211--252, 2015.

\bibitem{Sculley10}
D. Sculley.
\newblock Web-scale k-means clustering.
\newblock In Michael Rappa, Paul Jones, Juliana Freire, and Soumen Chakrabarti,
  editors, {\em WWW}, pages 1177--1178, 2010.

\bibitem{0001I20}
Tongzhou Wang and Phillip Isola.
\newblock Understanding contrastive representation learning through alignment
  and uniformity on the hypersphere.
\newblock In {\em ICML}, volume 119, pages 9929--9939, 2020.

\bibitem{WeinbergerS09}
Kilian~Q. Weinberger and Lawrence~K. Saul.
\newblock Distance metric learning for large margin nearest neighbor
  classification.
\newblock {\em J. Mach. Learn. Res.}, 10:207--244, 2009.

\bibitem{wu2019detectron2}
Yuxin Wu, Alexander Kirillov, Francisco Massa, Wan-Yen Lo, and Ross Girshick.
\newblock Detectron2.
\newblock \url{https://github.com/facebookresearch/detectron2}, 2019.

\bibitem{WuXYL18}
Zhirong Wu, Yuanjun Xiong, Stella~X. Yu, and Dahua Lin.
\newblock Unsupervised feature learning via non-parametric instance
  discrimination.
\newblock In {\em CVPR}, pages 3733--3742, 2018.

\bibitem{abs-1708-03888}
Yang You, Igor Gitman, and Boris Ginsburg.
\newblock Scaling {SGD} batch size to 32k for imagenet training.
\newblock {\em CoRR}, abs/1708.03888, 2017.

\bibitem{ZbontarJMLD21}
Jure Zbontar, Li Jing, Ishan Misra, Yann LeCun, and St{\'{e}}phane Deny.
\newblock Barlow twins: Self-supervised learning via redundancy reduction.
\newblock In Marina Meila and Tong Zhang, editors, {\em ICML}, volume 139 of
  {\em Proceedings of Machine Learning Research}, pages 12310--12320. {PMLR},
  2021.

\bibitem{ZhanX0OL20}
Xiaohang Zhan, Jiahao Xie, Ziwei Liu, Yew{-}Soon Ong, and Chen~Change Loy.
\newblock Online deep clustering for unsupervised representation learning.
\newblock In {\em CVPR}, 2020.

\bibitem{ZhuangZY19}
Chengxu Zhuang, Alex~Lin Zhai, and Daniel Yamins.
\newblock Local aggregation for unsupervised learning of visual embeddings.
\newblock In {\em ICCV}, pages 6001--6011, 2019.

\end{thebibliography}
}

\appendix

\section{Theoretical Analysis}
\subsection{Proof of Theorem~1}
\begin{proof}
Let the Lagrangian function at the $i$-th iteration be
\[\LL_i(\mu_i,\rho^{i-1}) = \sum_k s_{i,k}\mu_{i,k} + \sum_k\rho_k^{i-1}(\mu_{i,k}-\gamma_k/N)\]
where $\mu_i\in\R^{K}$,
and the solution for assignment is 
\begin{eqnarray}\label{eqs:solution}
\tilde{\mu}_{i,k} = \left\{\begin{array}{cc}1&k=\arg\max_k s_{i,k}+\rho_k^{i-1}\\0&o.w.\end{array}\right.
\end{eqnarray}
It is the maximal solution for the subproblem, and we have
\begin{eqnarray}\label{eqs:1}
\forall \mu_i,\quad \LL_i(\mu_i,\rho^{i-1})\leq \LL_i(\tilde{\mu}_i,\rho^{i-1})
\end{eqnarray}
where $\mu_i$ is an arbitrary assignment and $\tilde{\mu}_i$ is the assignment implied in Eqn.~\ref{eqs:solution}.
If fixing $\tilde{\mu}_i$ and assuming $\sum_k \gamma_k\leq N$, which is due to the fact that $\gamma_k$ is the lower-bound for each cluster size, we have the inequality for the arbitrary dual variables $\rho$ from the target convex domain as
\begin{align}\label{eqs:2}
&\LL_i(\tilde{\mu}_i,\rho^{i-1})-\LL_i(\tilde{\mu}_i,\rho)   = \sum_k(\rho_k^{i-1}-\rho_k)(\tilde{\mu}_{i,k}-\gamma_k/N) \nonumber\\
&\leq \frac{\|\rho^{i-1}-\rho\|_2^2-\|\rho^{i}-\rho \|_2^2}{2\eta} + \eta
\end{align}
Combining Eqns.~\ref{eqs:1} and \ref{eqs:2}, we have
\[\LL_i(\mu_i,\rho^{i-1})-\LL_i(\tilde{\mu}_i,\rho) \leq \frac{\|\rho^{i-1}-\rho\|_2^2-\|\rho^{i}-\rho \|_2^2}{2\eta} + \eta\]
With the constraint $\|\rho\|_1\leq \tau$ and adding $i$ from $1$ to $N$, we have
\[\sum_{i=1}^N \LL_i(\mu_i,\rho^{i-1})-\LL_i(\tilde{\mu}_i,\rho) \leq \frac{\tau^2}{2\eta}+\eta N\]
By setting $\eta = \frac{\tau}{\sqrt{2N}}$, it becomes
\[\sum_{i=1}^N \LL_i(\mu_i,\rho^{i-1})-\LL_i(\tilde{\mu}_i,\rho) \leq \tau\sqrt{2N}\]
Taking $\mu$ as the optimal solution for the original linear programming problem as $\mu^*$, we have
\begin{eqnarray*}
&&\R(\tilde{\mu}) + \sum_k\rho_k(\gamma_k-\sum_i\tilde{\mu}_{i,k})\\
&&\leq \sum_i \sum_k\rho_k^{i-1}(\gamma_k/N-\mu_{i,k}^*)+\tau\sqrt{2N}
\end{eqnarray*}
Let $\rho$ be the one-hot vector if there is violation.
\[\rho_k = \left\{\begin{array}{cc}\tau&k = \arg\max_k \gamma_k-\sum_i \tilde{\mu}_{i,k}\ and\ \V(\tilde{\mu})>0\\0&o.w.\end{array}\right.\]
Then, we can obtain the relationship between regret and violation as
\[\R(\tilde{\mu}) +\tau\V(\tilde{\mu})\leq \sum_i \sum_k\rho_k^{i-1}(\gamma_k/N-\mu_{i,k}^*)+\tau\sqrt{2N}\]
By assuming that the instances arrive in a stochastic order, we have $E[\gamma_k/N-\mu_{i,k}^*]\leq 0$ and the bound becomes
\begin{align*}
E[\R(\tilde{\mu})] \leq \tau\sqrt{2N};\quad \tau E[\V(\tilde{\mu})]\leq \tau \sqrt{2N} - E[\R(\tilde{\mu})]
\end{align*}
Now, we try to lower bound $\R(\tilde{\mu})$. The following analysis is for the case of $\V(\tilde{\mu})>0$. Since the violation is $\V(\tilde{\mu})$, we shrink the current solution $\tilde{\mu}$ by a factor of $\alpha=\min_k\frac{\gamma_k}{\gamma_k+K\V(\tilde{\mu})}$ such that there is no cluster with the number of instances more than $\gamma_k$ or there is at least $K\V(\tilde{\mu})$ unassigned instances. The shrunk solution with the re-assignment for the extra instances can be a feasible solution for the original assignment problem, so we have
\[\alpha \sum_i\sum_k s_{i,k} \tilde{\mu}_{i,k} \leq \mathrm{OPT}\]
where $\mathrm{OPT}$ denotes the optimal feasible result from $\mu^*$.
The lower-bound for $\R(\tilde{\mu})$ is
\begin{align*}
&E[\R(\tilde{\mu})]\geq E[(1-\frac{1}{\alpha})OPT]\geq -\frac{KE[\V(\tilde{\mu})]}{\min_k \gamma_k}\mathrm{OPT}
\end{align*}
Taking it back to the inequality for the violation and let $\tau$ be sufficiently large, the bound for the violation is obtained as
\begin{align*}
&E[\V(\tilde{\mu})] \leq \frac{1}{1 - \frac{K\mathrm{OPT}}{\tau\min_k \gamma_k}}  \sqrt{2N}
\end{align*}
\end{proof}

\subsection{Proof of Corollary~1}
\begin{proof}
Since $\{\x,C,\mu\}$ are sequentially updated, with $\LL'(\x,C, \mu)$ denoting the objective
\begin{scriptsize}
\begin{align}\label{eqs:uloss}
\min_{\x, C, \mu\in\Delta}  \sum_{i}\Big((K-1)\sum_{k=1}^K \mu_{i,k}\|\x_i-\cc_k\|_2^2-  \sum_{q=1}^K(1-\mu_{i,q})\|\x_i-\cc_q\|_2^2\Big)
\end{align}
\end{scriptsize}
we have $\LL'(\x^{t-1},C^{t-1},\mu^{t-1})\geq \LL'(\x^t,C^{t-1},\mu^{t-1})$ and $\LL'(\x^{t},C^{t-1},\mu^{t})\geq \LL'(\x^t,C^{t},\mu^{t})$. Therefore, the convergence for the bounded loss in Eqn.~\ref{eqs:uloss} can be guaranteed if $\LL'(\x^{t},C^{t-1},\mu^{t-1})\geq \LL'(\x^t,C^{t-1},\mu^{t})$. Since Theorem~1 indicates that $\mu^t$ is a near-optimal solution, it can reduce the loss effectively to make the inequality hold. Theoretically, we can keep $\mu^{t-1}$ when $\mu^t$ provides no loss reduction to guarantee the convergence.
\end{proof}

\section{Experiments}
\subsection{Implementation Details}
CoKe is learned with LARS optimizer~\cite{abs-1708-03888}, where weight decay is $10^{-6}$ and momentum is $0.9$. Batch size is set to $1,024$, where all experiments except the one with the multi-crop trick can be implemented on a server with $8$ GPUs and $16$G memory for each GPU. CoKe with two $224\times 224$ crops and six $96\times 96$ crops costs about $20$G memory on each GPU and is implemented on a server with $8$ GPUs and $32$G memory for each GPU. Learning rate is $1.6$ with cosine decay and the first $10$ epochs are used for warm-up. Batch normalization~\cite{IoffeS15} is synchronized across different GPUs as in~\cite{CaronMMGBJ20,ChenK0H20}. Augmentation is important for the performance of unsupervised representation learning~\cite{abs-2003-04297}, and we apply the same augmentation as in others~\cite{CaronMMGBJ20,ChenK0H20} that includes random crop, color jitter, random grayscale, Gaussian blur, and random horizontal flips. ResNet-50~\cite{HeZRS16} is adopted as the backbone and we apply a 2-layer MLP head to the backbone as suggested in~\cite{ChenK0H20,abs-2003-04297} for ablation experiments. The output dimension after MLP projection is $128$, which is also the same as benchmark methods~\cite{CaronMMGBJ20,ChenK0H20,abs-2003-04297}. 

To compare with state-of-the-art methods, we apply more sophisticated settings proposed by recent methods~\cite{GrillSATRBDPGAP20, ChenKSNH20}, e.g., $1000$ epoch training, $3$-layer projection MLP and $2$-layer prediction MLP. The training epoch for CoKe with multi-view is $800$. For Coke with two views, we set $\alpha=0.2$ and the ablation study for $\alpha$ can be found in the appendix. The temperature for CoKe with two/multi-view is reduced to $0.05$. Other settings, including batch size, dimension of representations, etc. remain the same. Following \cite{abs-2104-02057}, the linear classifier is optimized with SGD while the batch size and the number of epochs is $1,024$ and $90$, respectively. We conduct the ablation study for these additional components.

\subsection{Ablation study}
\subsubsection{Small Batch Training}
Since our objective for representation learning is a classification problem, it is insensitive to small batch size. To validate the claim, we have the experiments with the batch size of $\{256, 512, 1024\}$ in Table~\ref{tas:bs}. The learning rate for the batch size $256$ and $512$ is set to $0.8$ and $1.2$, respectively.
\begin{table}[!ht]
\centering
\begin{tabular}{|l|l|l|l|l|}\hline
Batch Size&256&512&1,024\\\hline
Acc\% &64.2&64.7&64.5 \\\hline
\end{tabular}
\caption{Comparison of different batch size.}\label{tas:bs}
\end{table}
We can observe from Table~\ref{tas:bs} that the performance of size $256$ is similar to that of $1,024$. It confirms that the proposed method is applicable with small batch size. Note that the ablation study has $200$ epochs for pre-training, and additional training epochs can further mitigate the gap as illustrated in SwAV~\cite{CaronMMGBJ20}.

\subsubsection{Single View with Moving Average}
Here, we investigate the effect of the proposed moving average strategy as a two-stage training scheme. To keep the label vector sparse, we fix the number of non-zero terms in a label vector to be $5$ in the second stage, where the performance is quite stable with other values in $\{10,20,30\}$. The sparse label will be further smoothed by a Softmax operator as 
\[\tilde{\mu}_{i,k} = \left\{\begin{array}{cc}\exp(\tilde{\mu}_{i,k}/\lambda')/Z&\tilde{\mu}_{i,k}>0\\0&\tilde{\mu}_{i,k}=0\end{array}\right.\]
where $Z = \sum_k I(\tilde{\mu}_{i,k}>0)\exp(\tilde{\mu}_{i,k}/\lambda')$ and $\lambda'=0.5$ in all experiments. We also update centers only after each epoch in the second stage as discussed in Sec.~3.2.2.

\begin{table}[!ht]
\centering
\begin{tabular}{|l|l|l|l|}\hline
$T'$&120&160&200\\\hline
Acc\% &64.3&65.8&65.3 \\\hline
\end{tabular}
\caption{Moving average with different $T'$.}\label{tas:twostage}
\end{table}

$T'$ is the number of epochs for the first stage and different settings of $T'$ is compared in Table~\ref{tas:twostage}. It can be observed that a single stage training strategy achieves $65.3\%$ accuracy, while smoothing the labels and centers in the last $40$ epochs can further improve the performance to $65.8\%$. It shows that the averaging strategy is effective for our framework. However, the performance will degrade if we begin moving average at an early stage as $T'=120$, which is due to that the model has not been trained sufficiently in the first stage. Given $T$, we will set $T'$ according to the ratio of $160/200$ for CoKe of single view.

\begin{figure*}[!ht]
\centering
\begin{minipage}{0.45\linewidth}
\centering
\includegraphics[height=1.35in]{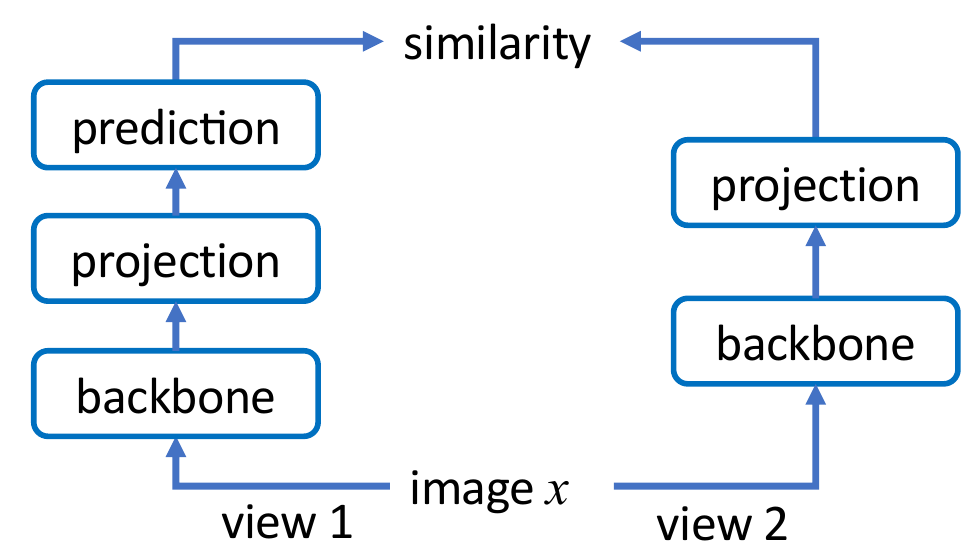}
\mbox{\footnotesize (a) Existing methods with prediction}
\end{minipage}
\begin{minipage}{0.45\linewidth}
\centering
\includegraphics[height=1.35in]{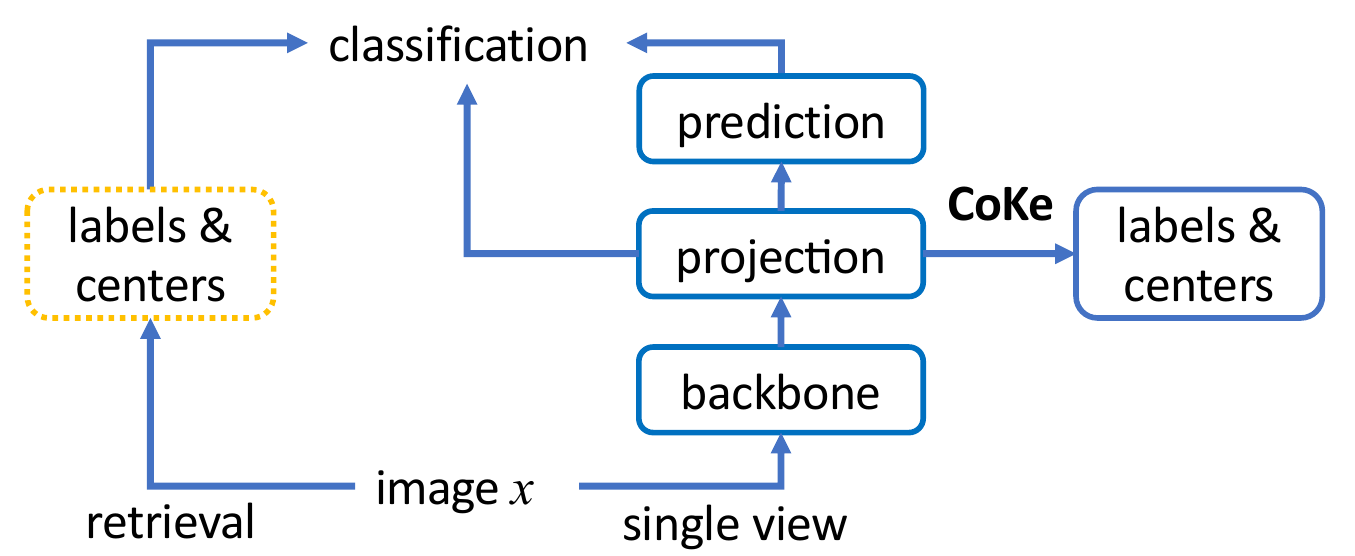}
\mbox{\footnotesize (b) CoKe with prediction}
\end{minipage}
\caption{Illustration of architecture with the additional prediction head. The yellow bounding box denotes the results from the last epoch.}\label{figs:pred}
\end{figure*}

\subsubsection{Optimization with Two Views}
Now we evaluate the model with sophisticated settings. When optimizing CoKe with two views, we keep the one-stage training scheme but improve pseudo labels as follows
\[\hat{\y}_{i:1}^{t} = \alpha\tilde{\y}_i^{t-1} + (1-\alpha)\p_{i:2}^{t-1}\]
where 
\[p_{i:j,q}^{t-1} = \frac{\exp(\x_i^{j\top} \cc_{q}^{t-1}/\lambda)}{\sum_{k=1}^K \exp(\x_i^{j\top} \cc_{k}^{t-1}/\lambda)}\]
The only parameter in the formulation is $\alpha$ that balances the one-hot label from the last epoch and soft label from the other view. The effect by varying $\alpha$ is summarized in Table~\ref{tas:alpha}. It demonstrates that a sufficiently large $\alpha$, which contains the information from the last epoch, is essential for improving the performance. We will fix $\alpha=0.2$ for rest experiments. 

\begin{table}[!ht]
\centering
\begin{tabular}{|l|l|l|l|l|}\hline
$\alpha$&0.1&0.2&0.3&0.4\\\hline
Acc\% &73.9&74.9&74.5&74.4 \\\hline
\end{tabular}
\caption{CoKe of two views with different $\alpha$.}\label{tas:alpha}
\end{table}

\subsubsection{Optimization with Prediction MLP}
We illustrate the different architectures with additional prediction head for existing methods~\cite{GrillSATRBDPGAP20, ChenKSNH20} and CoKe in Fig.~\ref{figs:pred}. Most of existing methods constrain that the representation after the prediction head is close to the representation of the other view after the projection head. Unlike those methods, CoKe tries to pull the representation of each instance to its corresponding cluster center. Therefore, both of representations after the projection MLP and prediction MLP can be leveraged for optimization as shown in Fig.~\ref{figs:pred} (b). Let $\ell_{\mathrm{pred}}$ and $\ell_{\mathrm{proj}}$ denote the classification loss for the representations from prediction and projection MLP, respectively. The final loss can be obtained as 
\[\ell = \beta \ell_{\mathrm{pred}} + (1-\beta) \ell_{\mathrm{proj}}\]
The effect of $\beta$ is summarized in Table~\ref{tas:beta} for CoKe with single view and two views. Note that the clustering phase including obtaining centers is applied to the representations after projection MLP only.

\begin{table}[!ht]
\centering
\begin{tabular}{|l|l|l|l|}\hline
$\beta$&0&0.5&1\\\hline
single-view &72.3 & 72.5&72.5\\\hline
two-view&74.4&74.9&73.3\\\hline
\end{tabular}
\caption{CoKe with different settings of prediction MLP.}\label{tas:beta}
\end{table}

From the comparison, it demonstrates that the additional prediction MLP is helpful for learning better representations. Interestingly, if optimizing the loss defined on representations from the prediction only, the performance of CoKe with two views can be degenerated. It may be due to that the dense soft label in two-view optimization is generated from the representation of the projection head. Without the corresponding loss, it is hard to optimize the prediction MLP solely.

\subsubsection{Long Training}
Finally, we compare the performance of $800$-epoch training to that of $1000$-epoch training in Table~\ref{tas:long}. Evidently, a longer training still can improve the performance slightly.

\begin{table}[!ht]
\centering
\begin{tabular}{|l|l|l|}\hline
\#epochs&800&1,000\\\hline
Acc\% &74.5&74.9\\\hline
\end{tabular}
\caption{CoKe of two views with different training epochs.}\label{tas:long}
\end{table}

\subsection{Comparison on Downstream Tasks}

After evaluating the performance on ImageNet, we apply the pre-trained models on downstream tasks for object detection, instance segmentation and classification. Four benchmark data sets are included for comparison. Concretely, we fine-tune Faster R-CNN~\cite{RenHG017} with R50-C4 as the backbone on PASCAL VOC~\cite{EveringhamGWWZ10} and Mask R-CNN~\cite{HeGDG17} with R50-FPN as the backbone and ``$1\times$'' training paradigm on COCO~\cite{LinMBHPRDZ14}. The codebase of MoCo~\footnote{https://github.com/facebookresearch/moco/tree/main/detection} with Detectron2~\cite{wu2019detectron2} is adopted. The standard fine-tuning procedure is applied for classification on CIFAR-10~\cite{krizhevsky2009learning} and CIFAR-100~\cite{krizhevsky2009learning}.

For object detection and instance segmentation, we follow the settings in MoCo~\cite{He0WXG20} for a fair comparison while only the learning rate is tuned. To obtain the optimal performance for each model, we search the learning rate in $[0.02,0.12]$ and $[0.01, 0.05]$ with a step size of $0.01$ for all methods on VOC and COCO, respectively. For classification, we search the learning rate in $\{1, 10^{-1}, 10^{-2}, 10^{-3}\}$ and weight decay in $\{10^{-5}, 10^{-6}, 0\}$, respectively. Besides, the learning rate for the last fully-connected layer is $10$ times larger than others since it is randomly initialized without pre-training. The best performance for baseline methods is reported. CoKe with two-view optimization is evaluated in this subsection.

\begin{table}[!ht]
\centering
\small
\begin{tabular}{|l|l|l|l|l|l|}\hline
&VOC&\multicolumn{2}{c|}{COCO}&C10&C100\\\hline
Methods&Ap$_{50}$&Ap$^{bb}$&Ap$^{mk}$&Acc\%&Acc\%\\\hline
Supervised&81.3&38.9&35.4 &97.3&86.6\\\hline
MoCo-v2&\underline{83.0}&39.6&35.9&97.9&86.1\\\hline
Barlow Twins&81.5&40.1&36.9&98.0&87.4 \\\hline
BYOL& 82.9&\underline{40.5}&36.9&\underline{98.1}&\underline{87.9}  \\\hline
SwAV$^*$&82.1&40.4&\underline{37.1}&97.7&87.5 \\\hline
DINO$^*$&82.0&40.2&36.8&97.7& 87.6\\\hline
CoKe&\underline{83.2}&\underline{40.9}&\underline{37.2}&\underline{98.2}&\underline{88.2}\\\hline
\end{tabular}
\caption{Comparison on downstream tasks. $^*$ denotes the usage of the multi-crop training trick. Top-2 best models are underlined.}\label{tas:ds}
\end{table}

Table~\ref{tas:ds} summarizes the standard metric on VOC, COCO and CIFAR. Explicitly, CoKe can outperform the supervised pre-trained model. It implies that an effective pre-trained model can be learned without supervision. Detailed reports on COCO can be found in Tables~\ref{tas:cocob} and~\ref{tas:cocom}.

\begin{table}[!ht]
\centering
\begin{tabular}{|l|l|l|l|}\hline
Methods&Ap$^{bb}$&Ap$^{bb}_{50}$&Ap$^{bb}_{75}$\\\hline
Supervised&38.9&59.6&42.7 \\\hline
MoCo-v2&39.6&60.5&43.4 \\\hline
Barlow Twins&40.1&61.6&43.9 \\\hline
BYOL&40.5&61.8&44.2   \\\hline
SwAV$^*$&40.4&61.8&44.0 \\\hline
DINO$^*$&40.2&61.7&43.8\\\hline
CoKe&40.9&62.3&44.7\\\hline
\end{tabular}
\caption{Comparison of object detection on COCO. }\label{tas:cocob}
\end{table}

\begin{table}[!ht]
\centering
\begin{tabular}{|l|l|l|l|}\hline
Methods&Ap$^{mk}$&Ap$^{mk}_{50}$&Ap$^{mk}_{75}$\\\hline
Supervised&35.4&56.5&38.1 \\\hline
MoCo-v2&35.9&57.4&38.4\\\hline
Barlow Twins&36.9&58.5&39.6 \\\hline
BYOL&36.9&58.6&39.5   \\\hline
SwAV$^*$&37.1&58.7&39.8 \\\hline
DINO$^*$&36.8&58.3&39.5\\\hline
CoKe&37.2&59.1&39.9\\\hline
\end{tabular}
\caption{Comparison of instance segmentation on COCO.}\label{tas:cocom}
\end{table}

\subsection{Comparison on Clustering}
As a deep clustering method, we compare CoKe to the benchmark clustering algorithms on CIFAR-10 and CIFAR-100~\cite{krizhevsky2009learning} in Tables~\ref{tas:cf10} and \ref{tas:cf20}, respectively. We follow the evaluation protocol in SCAN that trains models on training set and then evaluates the performance on test set with the prediction from the model directly. For CIFAR-100, 20 superclass labels are used for comparison. 

CoKe with two-view optimization is adopted for comparison. Compared with ImageNet, the resolution of images in CIFAR is only $32\times 32$. Therefore, we change the parameter of random crop from $[0.08,1]$ to $[0.3, 1]$ to keep the semantic information of the image. The model is optimized with SGD for $400$ epochs. The batch size is $128$ and the learning rate is $0.2$. Since CIFAR is a balanced data set, the lower-bound constraint is set to $0.9$ and the learning rate for the dual variables is $0.1$. Other settings remain the same and we have the same parameters for different data sets. To compare with SCAN, we have the same ResNet-18 as the backbone in CoKe. SCAN has $10$ clustering heads with the same number of clusters for multi-clustering. To have explicit multi-clustering, CoKe has $10$ clustering heads with different number of clusters. Concretely, we have $[10,100]$ with a step of $10$ for CIFAR-10 and $[20,200]$ with a step size of $20$ for CIFAR-100-20. The head with the target number of clusters is adopted for evaluation. The result averaged over $10$ trails is reported.

\begin{table}[!ht]
\centering
\begin{tabular}{|l|l|l|l|}\hline
\multirow{2}*{Methods}&\multicolumn{3}{c|}{CIFAR-10}\\\cline{2-4}
&ACC&NMI&ARI\\\hline
Supervised&93.8&86.2&87.0\\\hline
DeepCluster~\cite{CaronBJD18}&37.4&N/A&N/A\\\hline
IIC~\cite{JiVH19}&61.7&51.1&41.1\\\hline
Pretext~\cite{ChenK0H20}+k-means&65.9$\pm$5.7&59.8$\pm$2.0&50.9$\pm$3.7\\\hline
SCAN~\cite{GansbekeVGPG20} &81.8$\pm$0.3 & 71.2$\pm$0.4 & 66.5$\pm$0.4\\\hline
CoKe &85.7$\pm$0.2&76.6$\pm$0.3&73.2$\pm$0.4\\\hline
\end{tabular}
\caption{Comparison of clustering on CIFAR-10. SCAN is a two-stage method including pre-training and fine-tuning for clustering.}\label{tas:cf10}
\end{table}

\begin{table}[!ht]
\centering
\begin{tabular}{|l|l|l|l|}\hline
\multirow{2}*{Methods}&\multicolumn{3}{c|}{CIFAR-100-20}\\\cline{2-4}
&ACC&NMI&ARI\\\hline
Supervised&80.0&68.0&63.2\\\hline
DeepCluster~\cite{CaronBJD18}&18.9&N/A&N/A\\\hline
IIC~\cite{JiVH19}&25.7&22.5&11.7\\\hline
Pretext~\cite{ChenK0H20}+k-means&39.5$\pm$1.9&40.2$\pm$1.1&23.9$\pm$1.1\\\hline
SCAN~\cite{GansbekeVGPG20} &42.2$\pm$3.0 & 44.1$\pm$1.0 & 26.7$\pm$1.3 \\\hline
CoKe &49.7$\pm$0.7&49.1$\pm$0.4&33.5$\pm$0.4\\\hline
\end{tabular}
\caption{Comparison of clustering on CIFAR-100-20.}\label{tas:cf20}
\end{table}

\begin{figure}[!ht]
\centering
\begin{minipage}{\linewidth}
\centering
\includegraphics[height=0.31in]{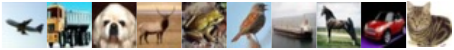}
\mbox{\footnotesize (a) CIFAR-10}
\end{minipage}

\begin{minipage}{\linewidth}
\centering
\includegraphics[height=0.6in]{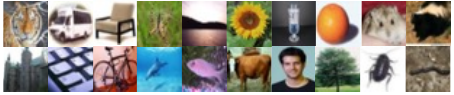}
\mbox{\footnotesize (b) CIFAR-100-20}
\end{minipage}
\caption{Exemplars obtained by CoKe on CIFAR.}\label{figs:cifa}
\end{figure}

Tables~\ref{tas:cf10} and \ref{tas:cf20} show that CoKe as an end-to-end clustering framework can achieve the superior performance without any fine-tuning. On the contrary, SCAN has a two-stage training strategy that learns representations in the first stage with instance discrimination and fine-tunes the model for clustering with different objectives and augmentations in the second stage. Therefore, the representation from the first stage may degenerate the performance of clustering. Finally, Fig.~\ref{figs:cifa} shows the exemplars that are close to cluster centers from CoKe on CIFAR. We can find that CoKe can recover the exact classes on CIFAR-10, which confirms the effectiveness of CoKe for clustering.

\end{document}